\documentclass{article}

\usepackage{arxiv}

\usepackage[utf8]{inputenc} 
\DeclareUnicodeCharacter{03C9}{$\omega$} 
\usepackage[T1]{fontenc}    
\usepackage{hyperref}       
\usepackage{url}            
\usepackage{booktabs}       
\usepackage{amsfonts}       
\usepackage{nicefrac}       
\usepackage{microtype}      
\usepackage{lipsum}

\usepackage{amsmath,amsfonts}
\usepackage{array}
\usepackage{textcomp}
\usepackage{stfloats}
\usepackage{url}
\usepackage{verbatim}
\usepackage{graphicx}
\usepackage{cite}

\usepackage{amsfonts}
\usepackage{amsmath}
\usepackage{subfig}
\usepackage{amssymb}
\usepackage{mathtools}
\usepackage{xcolor}
\usepackage{caption}
\usepackage{float}
\usepackage[ruled,vlined]{algorithm2e}
\usepackage{stmaryrd}
\usepackage{enumitem}
\usepackage{subcaption}
\usepackage{hyperref}
\usepackage{wrapfig}

\newcommand{\col}{\textsf{ col}}

\newcommand{\R}{{\mathbb{R}}}

\newcommand{\N}{{\mathbb{N}}}

\newcommand{\tx}{{t_{\times}}}

\newtheorem{theorem}{Theorem}[section]

\newtheorem{assumption}{Assumption}

\newtheorem{remark}[theorem]{Remark}
\newtheorem{problem}[theorem]{Problem}
\newtheorem{proof}[theorem]{Proof}

\title{Prescribed Performance Control of Unknown Euler–Lagrange Systems Under Input Constraints
\thanks{ This work was supported in part by the SERB Start-Up Research Grant; in part by the ARTPARK. The work of Ratnangshu Das was supported by the Prime Minister’s Research Fellowship from the Ministry of Education, Government of India.}
}

\author{
 Ratnangshu Das \\
  Centre for Cyber-Physical Systems\\
  IISc, Bengaluru, India\\
  \texttt{ratnangshud@iisc.ac.in} \\
  \And
 Pushpak Jagtap \\
  Centre for Cyber-Physical Systems\\
  IISc, Bengaluru, India\\
  \texttt{pushpak@iisc.ac.in} \\
}

\begin{document}

\maketitle

\begin{abstract}
In this paper, we present a prescribed performance control framework for trajectory tracking in Euler–Lagrange systems with unknown dynamics and prescribed input constraints. The proposed approach enforces hard funnel constraints, meaning that the prescribed performance bounds must not be violated during operation. We derive feasibility conditions that guarantee the tracking error evolves within these predefined funnels while ensuring bounded control inputs. To handle situations where the feasibility conditions are not satisfied, we introduce two approximation-free control strategies: one that actively drives the error back toward the funnel and another that prioritizes safety by preventing further deviation. The effectiveness and robustness of the proposed method are demonstrated through simulation studies and hardware experiments, highlighting its suitability for real-world robotic systems operating under strict input limits.
\end{abstract}


\section{Introduction} \label{sec:intro}
Autonomous systems such as self-driving cars, robots, and manipulators have generated a growing demand for tracking controllers that come with formal guarantees \cite{formalintro}. In practice, this problem is challenging due to unknown or partially known dynamics and external disturbances \cite{unknown1}, and becomes particularly significant in safety-critical Euler–Lagrange systems under strict input constraints. 

Classical tracking controllers such as sliding mode control \cite{SMC1}, and model predictive control \cite{MPC2} have been widely used in robotics. While effective in many settings, these methods either rely heavily on gain tuning, suffer from chattering or computational complexity, or require accurate system models. Learning-based approaches \cite{RLtracking2} can handle uncertainty, but often introduce additional complexity and lack formal safety guarantees.

Prescribed performance control (PPC) \cite{PPC1, PPC2} has emerged as an attractive alternative for systems with unknown dynamics. By constraining the tracking error within predefined time-varying funnels, these methods provide explicit guarantees on transient and steady-state performance using simple closed-form control laws. PPC has been successfully applied to a wide range of problems, including multi-agent systems \cite{Funnel_MAS2}, stochastic systems \cite{jagtap2021distributed}, and temporal logic tasks \cite{das2025spatiotemporalomega}. Unlike classical PD/PID controllers, they explicitly provide formal bounds on tracking accuracy and constraint satisfaction, while avoiding the need for extensive gain tuning.

A key limitation of standard PPC, however, is the possibility of unbounded control inputs as the error approaches the funnel boundary. This limits its applicability in real world systems with hard input constraints. Existing input-constrained PPC methods mainly address linear or SISO systems \cite{Funnel_Input_Constraint_Linear, mishra2024approximation} and therefore do not readily extend to nonlinear multi-DOF EL dynamics. Alternatively, \cite{Funnel_Input_Constraint_Berger} handles input saturation by dynamically widening the funnel boundaries. While effective for feasibility, relaxing the funnels compromises the prescribed performance guarantees and is unsuitable in safety-critical settings where funnels represent hard constraints (e.g., physical obstacles or joint limits) that cannot be relaxed.

In this paper, we propose a prescribed performance controller for Euler–Lagrange systems with input constraints. The proposed method preserves the original prescribed performance boundaries as hard funnel constraints and explicitly accounts for input limits through feasibility conditions. When feasibility conditions are violated, we introduce two alternative bounded control strategies: (i) an error correction strategy that actively drives the system back into the funnel, and (ii) an error containment strategy that prioritizes safety by preventing further deviation. The results are validated through simulations and hardware experiments on robotic platforms, including mobile robots and manipulators.

\begin{figure}[!t]
\centering
\subfloat[]{\includegraphics[height=2.1in]{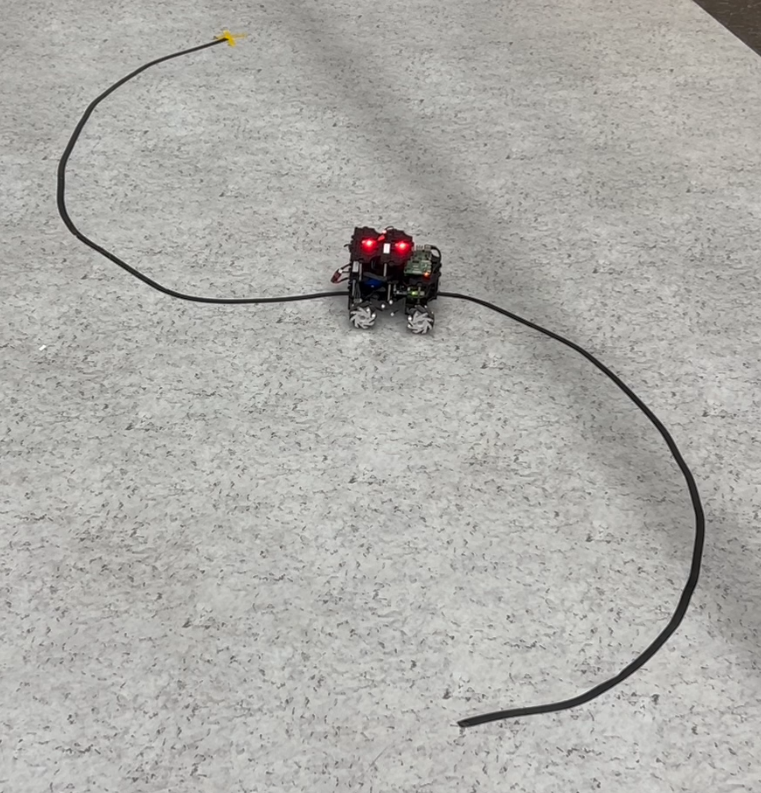}%
\label{fig:omni_rw}}
\hfill
\subfloat[]{\includegraphics[height=2.1in]{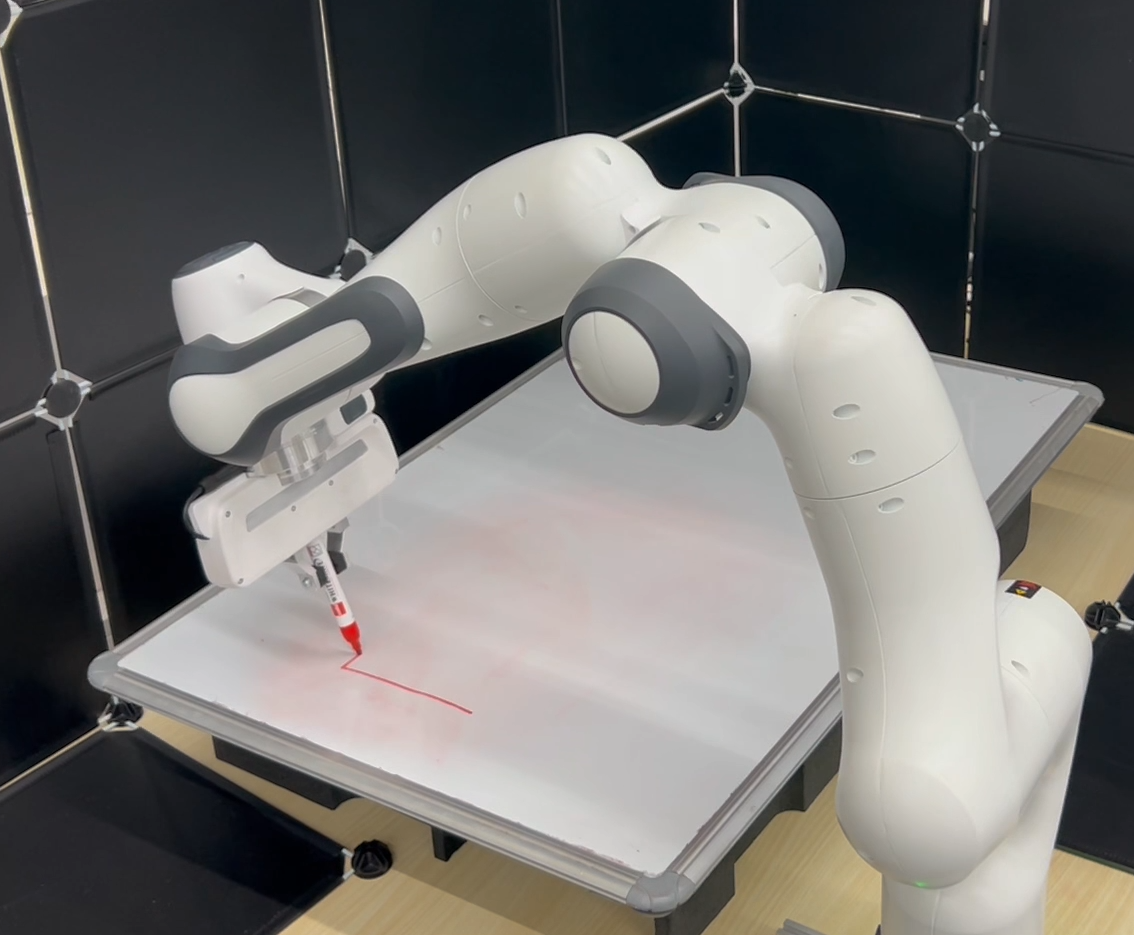}%
\label{fig:franka}}
\caption{(a) Omnidirectional mobile-robot. (b) FRANKA RESEARCH 3.}
\label{fig:robots}
\end{figure}

\section{Preliminaries and Problem Formulation} \label{sec:prob}

\subsection{Notations}
The symbols $\N$, $ \R$, $\R^+$, and $\R_0^+ $ denote the set of natural, real, positive real, and nonnegative real numbers, respectively. 
We use $ \R^{n\times m} $ to denote a vector space of real matrices with $ n $ rows and $ m $ columns. To represent a set of column vectors with $n$ rows, we use $ \R^{n}$.
To denote a vector $x \in \R^{n}$ with entries $x_1, \ldots, x_n$, we use $\col(x_1, \ldots, x_n)$, where $x_i \in \R, i \in [1;n]$ denotes the $i$-th element of the vector $x\in\R^n$. 
A diagonal matrix in $\R^{n\times n}$ with diagonal entries $d_1,\ldots, d_n$ is denoted by $\textsf{diag}(d_1,\ldots, d_n)$.
Given a vector $x \in \R^n$, we represent the element-wise absolute value using $|x| := \col(|x_1|, \ldots, |x_n|)$ and the Euclidean norm using $\|x\|$. 
For $a,b\in\R$ and $a< b$, we use $(a,b)$ to represent an open interval in $\R$. For $a,b\in\N$ and $a\leq b$, we use $[a;b]$ to denote a close interval in $\N$. 
For $x, y \in \R^n$, the vector inequalities, $x \preceq y$ (and $x \succeq y$) represents $x_i \leq y_i$ (and $x_i \geq y_i$), $\forall i \in [1;n]$.
We use $I_n$ and $\textbf{0}_{n\times m}$ to denote identity matrix in $\R^{n\times n}$ and zero matrix in $\R^{n\times m}$, respectively. 
$x \uparrow (\downarrow) \ a$ indicates $x$ approaches $a$ from the left (right) side. 

\subsection{System Definition}
Consider an Euler-Lagrange (EL) system $\mathcal{S}$ described by
\begin{align}
    \mathcal{S}: M(x)\Ddot{x} + V(x,\dot{x}) + G(x) = \tau + d(t), \label{eqn:sysdyn}
\end{align}
where $x(t) = [x_1(t), \ldots, x_n(t)]^\top \in X \subset \mathbb{R}^n$ is the system configuration, $\tau(t) \in \mathbb{R}^n$ is the control input, and $d(t) \in \mathbb{D} \subset \R^n$ is an unknown disturbance. The terms $M(x) \in \R^{n \times n}$, $V(x,\dot{x}) \in \R^n$, and $G(x) \in \R^n$ denote the inertia, Coriolis/centrifugal, and gravity components, respectively. For brevity, functional arguments are omitted, and the terms are denoted as $M,V,G$ and $d$.

\begin{assumption}
    The mass matrix $M$, the Coriolis and centrifugal terms $V$, the gravity vector $G$ and the external disturbance $d$ are all assumed to be unknown.
\end{assumption}

For the EL system $\mathcal{S}$, assume the control input is bounded as $|\tau(t)| \preceq \overline{\tau},$ for all $t \in \R_0^+$. This implies corresponding bounds on the system dynamics \cite[Chapter~2]{ELbook}. Although the disturbance $d$ and the system parameters $M$, $V$, and $G$ are unknown, their boundedness can be used for control design. Accordingly, we introduce the following assumptions.

\begin{assumption}\label{assum_d}
    The external disturbance $d$ satisfies $-\overline{d} \preceq d \preceq \overline{d},$ for all $t \in \R_0^+$, where $\overline{d} \in \R^n$ is a known bound.
\end{assumption}

\begin{assumption}\label{assum_tau}
    Given the control bound $\overline{\tau}$, there exists a positive constant $\underline{m} \in \R$, such that $\underline{m}\overline{\tau} \preceq M^{-1}\overline{\tau}$.
\end{assumption}

\begin{assumption}\label{assum_V}
The Coriolis and centrifugal terms $V$ and the gravity vector $G$ satisfy $\underline{V}_M \preceq V_M \preceq \overline{V}_M$, where $V_M := -M^{-1}(V+G)$ and $\underline{V}_M, \overline{V}_M \in \R^n$.
\end{assumption}

\begin{assumption}\label{assum_Md}
    The inverse of the mass matrix scales the disturbance as $-\|M^{-1}\| \overline{d} \preceq M^{-1}d \preceq \|M^{-1}\| \overline{d}$. This implies, there exists $\underline{m}_i \in \R^+$, such that $-\underline{m}_i \overline{d} \preceq M^{-1}d \preceq \underline{m}_i \overline{d}$.
\end{assumption}

The Assumptions \ref{assum_d}-\ref{assum_Md} provide the bounds on various system parameters, which are utilized for establishing the feasibility conditions in Section \ref{sec:feas}.

\subsection{Problem Statement}
The goal of this work is to design a feedback control law $\tau(t)$ such that the trajectory $x(t)$ of the unknown EL system~\eqref{eqn:sysdyn} tracks a given reference trajectory $x_{ref}(t) \in \R^n$, for all $t \in \R_0^+$, while respecting the input constraints $|\tau(t)| \preceq \overline{\tau}$. 

To ensure that the tracking problem is well posed, we make the following assumption on the reference trajectory $x_{ref}(t)$.
\begin{assumption}\label{assum_xref}
    The reference trajectory, $x_{ref}(t)$, is continuously differentiable, and its velocity is bounded, i.e., there exists $\overline{v}_r \in \R_0^+$ such that $|\dot{x}_{ref}(t)| \preceq \overline{v}_r$ for all $t \in \R_0^+$.
\end{assumption}

The control problem is formally stated as follows:
\begin{problem}\label{prob}
    Given the EL system~\eqref{eqn:sysdyn} and a reference trajectory $x_{ref}(t)$, satisfying Assumptions~\ref{assum_d}-\ref{assum_xref}, design a model-free feedback control law $\tau(t)$, such that: 
    \begin{itemize}
        \item[(i)] $x(t)$ tracks $x_{ref}(t)$ for all $t \in \R_0^+$, and
        \item[(ii)] the input constraint $|\tau(t)| \preceq \overline{\tau}$ is satisfied for all $t \in \R_0^+$.
    \end{itemize}
\end{problem}

\section{Controller Design and Theoretical Analysis}\label{sec:con}

First, a virtual velocity input is introduced to enforce the tracking specification, which is then extended to an acceleration-level control formulation.

In this section, we derive the control law formulated to address Problem~\ref{prob}. The proposed controller is developed using a systematic two-step procedure inspired by a backstepping-like design approach, similar to that described in \cite{PPCfeedback} and \cite{hard_soft}. First, a virtual velocity input is introduced to enforce the tracking specification, which is then extended to an acceleration-level control formulation. 
For convenience, the EL system~\eqref{eqn:sysdyn} is rewritten as
\begin{subequations}
    \begin{align}
    \dot{x} &= v, \label{eqn:sysDyn_vel}   \\
    \dot{v} &= M(x)^{-1} \left(-V(x, v) - G(x) + \tau + d \right) = V_M(x,v) + M(x)^{-1}\tau + M(x)^{-1}d, \label{eqn:sysDyn_acc}
    \end{align}
\end{subequations}
where $V_M(x,v) = -M(x)^{-1}(V(x,v)+G(x))$.

\subsection{Stage I}
Given a reference signal $x_{ref}(t)$, define the tracking error:
$$e_x(t) = x(t) - x_{ref}(t).$$ 
To enforce prescribed performance, we introduce exponentially decaying funnel constraints $\rho_x: \R_0^+ \rightarrow \R^n$:
\begin{equation}\label{eqn:rhox}
    \rho_x(t) = e^{-\mu_x t}(p_x-q_x) + q_x,    
\end{equation}
where $p_x \in \R^n$ specifies the initial funnel width with $|e_x(0)| \preceq p_x$, $q_x \in \R^n$ defines the steady-state bound with $\textbf{0}^{n \times 1} \prec q_x \prec p_x$, and $\mu_x \succeq \textbf{0}^{n \times n}$ is a diagonal matrix that determines the decay rate.
We enforce tracking by constraining the tracking error $e_x(t)$ to remain within the funnel for all $t \in \R_0^+$:
\begin{equation}\label{eqn:fun}
    -\rho_x(t) \prec e_x(t) \prec \rho_x(t). 
\end{equation}
To help with control design, define the normalized error:
\begin{equation}\label{eqn:norm_err}
    \varepsilon_x(t) = \text{diag}(\rho_{x})^{-1}e_x(t).    
\end{equation}
The velocity-level control input $v_r(t)$ is then formulated as: 
\begin{equation} \label{eqn:velcon}
    v_r(t) = -\overline{v}\text{diag}(\Psi(\varepsilon_x)),
\end{equation}
where the map $\Psi$ is the bounded transformation defined in Section~\ref{sec:clamp}, and $\overline{v}\in\mathbb{R}^n$ is the maximum allowable velocity.

This ensures that the tracking error remains constrained within the prescribed funnels, while respecting input bounds. 

\subsection{Stage II}
To smoothly track the reference velocity $v_{r}(t)$ from Stage I, as given in \eqref{eqn:velcon}, we now regulate the velocity error $e_v(t)$:
$$e_v(t) = v(t) - v_{r}(t).$$ 
To bound this error, we introduce exponentially decaying funnel constraints $\rho_v: \R_0^+ \rightarrow \R^n$, given by:
\begin{equation}\label{eqn:rhov}
    \rho_v(t) = e^{-\mu_v t}(p_v-q_v) + q_v,
\end{equation}
where $p_v \in \R^n$ is the initial funnel width, with $|e_v(0)| \preceq p_v$, $q_v \in \R^n$ is the steady state limit, with $0^{n \times 1} \prec q_v \prec p_v$, and $\mu_v \succeq \textbf{0}^{n \times n}$ determines the funnel decay rate.
The velocity error $e_v$ is constrained to remain within the funnel: 
\begin{equation}\label{eqn:fun2}
    -\rho_v(t) \prec e_v(t) \prec \rho_v(t).    
\end{equation}
Define the normalized velocity error $\varepsilon_v(t)$:
\begin{equation}\label{eqn:norm_err2}
    \varepsilon_v(t) = \text{diag}(\rho_{v})^{-1}e_v(t).
\end{equation}
The acceleration-level control input $\tau(t)$ is formulated as:
\begin{equation}\label{eqn:con}
    \tau(t) = -\overline{\tau}\text{diag}(\Psi(\varepsilon_v)),
\end{equation}
where $\Psi$ is the same bounded transformation function in Section~\ref{sec:clamp}, and $\overline{\tau} \in \R^n$ is the maximum permissible torque.

Prescribed transient and steady-state performance is achieved by keeping the position and velocity errors within their funnels~\eqref{eqn:fun} and~\eqref{eqn:fun2}, which is equivalent to maintaining the normalized errors within $(-1,1)^n$. Unlike standard PPC, the proposed law also enforces input constraints by introducing bounded transformation functions.

\subsection{{Bounded Transformation Function}} \label{sec:clamp}
The bounded transformation function $\Psi:\R^n \rightarrow \R^n$ is a continuously differentiable map that ensures that the control remains bounded within the control limits while maintaining the desired behavior. In the paper, we introduce two categories of bounded transformation functions.
\subsubsection{{Saturation Transformation Function}}
\begin{figure}[t]
\centering
\subfloat[Saturation Function]{%
\includegraphics[width=0.43\textwidth]{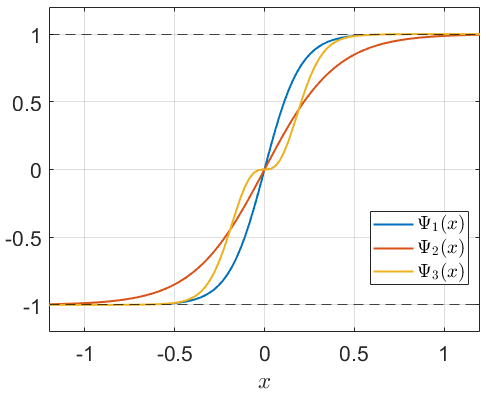}%
\label{fig:sat}}
\hfill
\subfloat[Zeroing Function]{%
\includegraphics[width=0.43\textwidth]{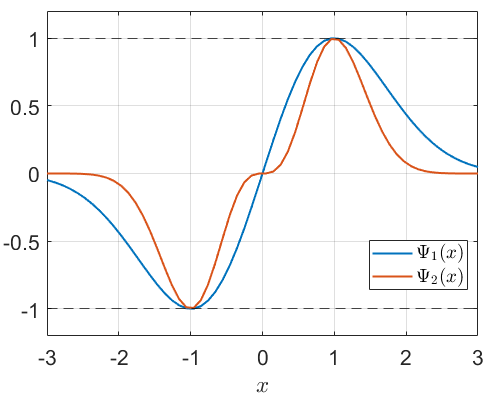}%
\label{fig:zeroing}}
\caption{Bounded Transformation Functions.}
\label{fig:clamp}
\end{figure}

We define this as $\Psi(s) = [\Psi_1(s_1), \ldots, \Psi_n(s_n)]^\top$, where for all $i = [1;n]$:
$$\Psi_i(s_i) = 
\begin{cases} 
  -1, & s_i \in (-\infty,-1], \\
  0, & s_i = 0, \\
  1, & s_i \in [1,\infty),
\end{cases}
$$
and $\Psi_i(\cdot)$ is nondecreasing. Examples are shown in 
Figure \ref{fig:sat}.
As $s_i$ extends beyond $1$ and $-1$, $\Psi_i(s_i)$ \textbf{saturates} at $1$ and $-1$, respectively. This ensures that even if the tracking error exceeds the funnel boudaries due to increased disturbances beyond feasible limits, the error is driven towards the funnel, enhancing the system's robustness against external disturbances.

\subsubsection{{Zeroing Transformation Function}}
We define this as $\Psi(s) = [\Psi_1(s_1), \ldots, \Psi_n(s_n)]^\top$, where for all $i = [1;n]$:
$$\Psi_i(s_i) = 
\begin{cases} 
  -1, & s_i \!=\! -1, \\
  0, & s_i \!=\! 0, \\
  1, & s_i \!=\! 1,
\end{cases}
\lim_{s_i \rightarrow \infty} \Psi_i(s_i) \!=\! 0, 
\lim_{s_i \rightarrow -\infty} \Psi_i(s_i) \!=\! 0,$$
and $\Psi_i(s_i)$ should be non-decreasing for all $s_i \in (-1,1)$.
Figure \ref{fig:zeroing} depicts a few examples of zeroing function.

Outside the interval $[-1,1]$, $\Psi_i(s_i)$ \textbf{decays} to $0$, which reduces the control effort when large deviations occur. This provides a safety-oriented response by preventing potentially aggressive control actions in unexpected operating conditions.

\begin{remark}
Classical PPC \cite{PPC1,PPC2} is not well defined when the tracking error leaves the prescribed funnel, which may lead to unbounded or undefined inputs. The bounded transformation functions introduced here resolve this issue by ensuring that the control signal remains bounded for all error values. Unlike standard input clamping or anti-windup schemes that modify the control after computation, the bounded transformation is embedded directly in the controller design. As a result, it guarantees bounded inputs by construction while provably preserving the prescribed performance guarantees.
\end{remark}

Now, we introduce the feasibility conditions, under which we proceed to guarantee that the tracking error remains constrained within the funnel bounds.

\subsection{Feasibility Condition}\label{sec:feas}
Maintaining the tracking error within the funnels guarantees desired transient and steady-state performance, while input constraints address practical concerns like actuator safety and control effort minimization. This introduces a trade-off between performance and available actuation. To balance this trade-off, we derive the following feasibility conditions.
\subsubsection{Stage I}
Given funnel constraints $\rho_x(t)$ in \eqref{eqn:rhox}, the maximum permissible velocity $\overline{v}$ must satisfy:
\begin{equation}\label{eqn:feas1}
    \overline{v} \succeq \mu_x(p_x-q_x) + \overline{v}_r + p_v.
\end{equation}

\subsubsection{Stage II}
For the funnel $\rho_v(t)$ in \eqref{eqn:rhov}, and the dynamics in~\eqref{eqn:sysdyn} with Assumptions~\ref{assum_d}-\ref{assum_Md}, the torque limit $\overline{\tau}$ must satisfy:
\begin{equation}\label{eqn:feas2}
    \overline{\tau} \succeq \frac{1}{\underline{m}} \left( \max(-\underline{V_M}, \overline{V_M}) + \underline{m}_i\overline{d} + \mu_v(p_v-q_v) + \overline{a}_r \right),
\end{equation}
where $\overline{a}_r \in \mathbb{R}^n$ is an upper bound on $|\dot{v}_r|$. For example, if $\Psi_i(s) := \tanh(5s)$, then one may choose $\overline{a}_r = 5 \overline{v}$.

\subsection{Tracking Error Analysis}
The theorem formally summarizes the approximation-free closed-form controller proposed in this paper.

\begin{theorem}\label{thm:bdcontrol}
    Consider the Euler–Lagrange system~\eqref{eqn:sysdyn} satisfying Assumptions~\ref{assum_d}-\ref{assum_Md}, and a reference trajectory under Assumption~\ref{assum_xref}. If the initial conditions satisfy $|e_x(0)| \!\prec\! p_x$ and $|e_v(0)| \!\prec\! p_v$, and the feasibility conditions~\eqref{eqn:feas1}-\eqref{eqn:feas2} hold, then the control law $\tau(t)$ in~\eqref{eqn:con} guarantees that for all $t \in \R_0^+$
    \begin{itemize}
        \item[(i)] Funnel invariance:
        $|e_x(t)| \!\prec\! \rho_x(t) \text{ and } |e_v(t)| \!\prec\! \rho_v(t),$
        \item[(ii)] Bounded Signals:
        $|\tau(t)| \preceq \overline{\tau} \text{ and } |v(t)| \preceq \overline{v}.$
    \end{itemize}
\end{theorem}

\begin{figure}[!t]
\centering
\subfloat[Case 1]{%
\includegraphics[width=0.43\textwidth]{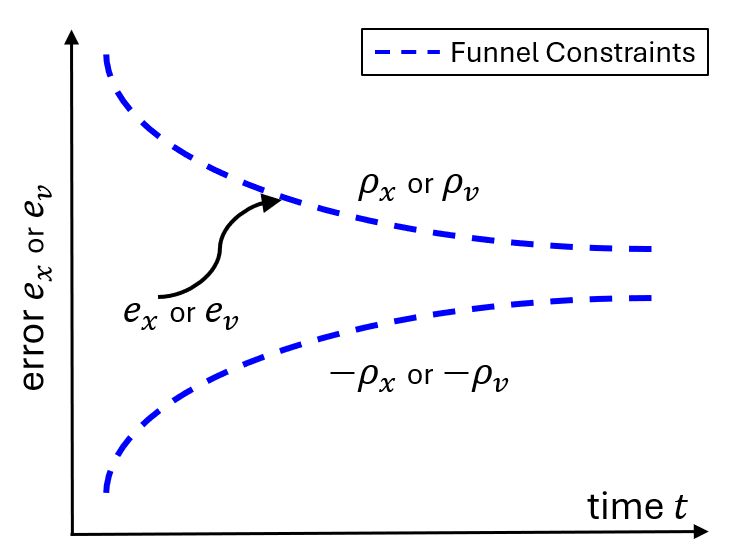}%
\label{fig:case1}}
\hfill
\subfloat[Case 2]{%
\includegraphics[width=0.43\textwidth]{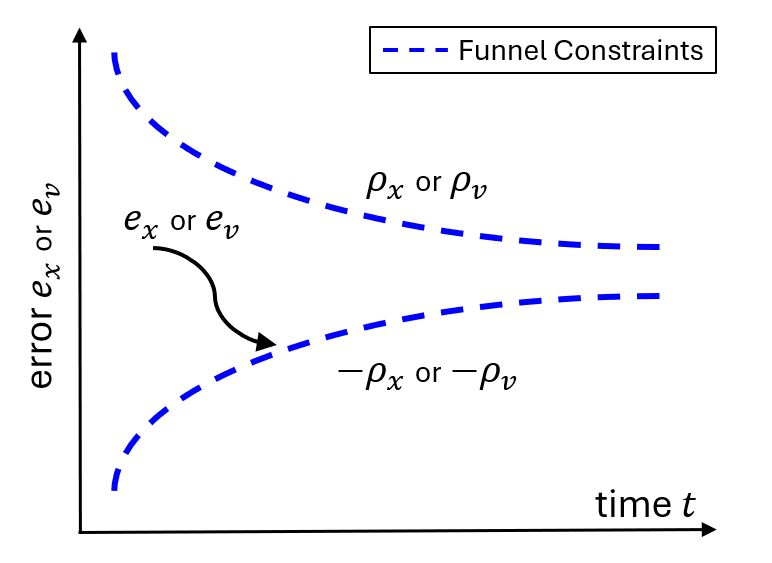}%
\label{fig:case2}}
\caption{Funnel constraints.}
\label{fig:case}
\end{figure}

\begin{proof}
    The proof is divided into two stages:

\textbf{Stage I: Invariance of the $x$-funnel.}
We use contradiction to show that the reference velocity $v_r(t)$ in~\eqref{eqn:velcon} keeps the position error inside the funnel $[-\rho_x(t),\rho_x(t)]$ as in Equation~\eqref{eqn:fun}.

Let $\tx$ be the first time at which
\eqref{eqn:fun} is violated. Then,
\begin{align}\label{Eq:inqe_tx}
    &\exists i \in [1;n], e_{x,i}(\tx) \leq -\rho_{x,i}(\tx) \text{ or } e_{x,i}(\tx) \geq \rho_{x,i}(\tx) \notag \\
    &-\rho_{x,i}(t) < e_{x,i}(t) < \rho_{x,i}(t), \forall (t,i) \in [0, \tx) \times [1;n].
\end{align}
We will consider the following two cases for $t \in [0,\tx)$.

\textbf{Case I.} For some $i \in [1;n]$, $e_{x,i}(t)$ approaches the upper funnel constraint (Figure \ref{fig:case1}), i.e., $e_{x,i}(t) \rightarrow \rho_{x,i}(t) \implies e_{x,i}(t) - \rho_{x,i}(t) =: \overline{\alpha}_{x,i} \rightarrow 0$. Following \eqref{Eq:inqe_tx}, we have:
\begin{align*}
    &e_{x,i}(t) < \rho_{x,i}(t) \implies \overline{\alpha}_{x,i} \uparrow 0 \implies \lim_{\overline{\alpha}_{x,i} \uparrow 0} \frac{d}{dt} \overline{\alpha}_{x,i} > 0 \\
    \implies &\lim_{\overline{\alpha}_{x,i} \uparrow 0} \dot{e}_{x,i}(t) > \lim_{\overline{\alpha}_{x,i} \uparrow 0} \dot{\rho}_{x,i}(t) > -\mu_{x,i}(p_{x,i}-q_{x,i}) \\
    \implies &\lim_{\overline{\alpha}_{x,i} \uparrow 0} \dot{x}_{i}(t) > -\mu_{x,i}(p_{x,i}-q_{x,i}) + \dot{x}_{ref}(t).
\end{align*}
Therefore, there exists $i \in [1;n]$, such that 
\begin{gather}\label{eqn:dx_b1}
    \lim_{\overline{\alpha}_{x,i} \uparrow 0} \dot{x}_{i}(t) > -\mu_{x,i}(p_{x,i}-q_{x,i}) - \overline{v}_{r,i}.
\end{gather}
Since, $\lim_{\overline{\alpha}_{x,i} \uparrow 0} \varepsilon_{x,i}(t) = 1$ we obtain $\lim_{\overline{\alpha}_{x,i} \uparrow 0} v_{r,i}(t) = -\overline{v}_{i}$. 
Now, the actual velocity $\dot{x}_i(t) = v_{r,i}(t) + e_{v,i}(t)$, where $e_{v,i}(t) < p_{v,i}$ (we ensure this in Stage II), with the feasibility condition~\eqref{eqn:feas1} gives:
\begin{align}\label{eqn:dx_b1c}
    \lim_{\overline{\alpha}_{x,i} \uparrow 0} \dot{x}_{i}(t) \leq -\overline{v}_{i} + p_{v,i} \leq -\mu_{x,i}(p_{x,i}-q_{x,i}) - \overline{v}_{r,i},
\end{align}
which contradicts \eqref{eqn:dx_b1}.

\textbf{Case II.} For some $i \in [1;n]$, $e_{x,i}(t)$ approaches the lower funnel constraint (Figure \ref{fig:case2}), i.e., $e_{x,i}(t) \rightarrow -\rho_{x,i}(t) \implies e_{x,i}(t)+\rho_{x,i}(t) =: \underline{\alpha}_{x,i} \rightarrow 0$. Following \eqref{Eq:inqe_tx}, we have:
\begin{align*}
    &e_{x,i}(t) > -\rho_{x,i}(t) \implies \underline{\alpha}_{x,i} \downarrow 0 \implies \lim_{\underline{\alpha}_{x,i} \downarrow 0} \frac{d}{dt} \underline{\alpha}_{x,i} < 0 \\
    \implies &\lim_{\underline{\alpha}_{x,i} \downarrow 0} \dot{e}_{x,i}(t) < \lim_{\underline{\alpha}_{x,i} \downarrow 0} -\dot{\rho}_{x,i}(t) < \mu_{x,i}(p_{x,i}-q_{x,i}) \\
    \implies &\lim_{\underline{\alpha}_{x,i} \downarrow 0} \dot{x}_{i}(t) < \mu_{x,i}(p_{x,i}-q_{x,i}) + \dot{x}_{ref}(t).
\end{align*}
Therefore, there exists $i \in [1;n]$, such that
\begin{gather}\label{eqn:dx_b2}
    \lim_{\overline{\alpha}_{x,i} \downarrow 0} \dot{x}_{i}(t) < \mu_{x,i}(p_{x,i}-q_{x,i}) + \overline{v}_{r,i}.
\end{gather}
Since, $\lim_{\underline{\alpha}_{x,i} \uparrow 0} \varepsilon_{x,i}(t) = -1$ we obtain $\lim_{\underline{\alpha}_{x,i} \uparrow 0} v_{r,i}(t) = \overline{v}_{i}$. 
Now, the actual velocity $\dot{x}_i(t) = v_{r,i}(t) + e_{v,i}(t)$, where $e_{v,i}(t) > -p_{v,i}$ (we ensure this in Stage II), with the feasibility condition~\eqref{eqn:feas1} gives:
\begin{align}\label{eqn:dx_b2c}
    \lim_{\underline{\alpha}_{x,i} \downarrow 0} \dot{x}_{i}(t) \geq \overline{v}_i - p_{v,i} \geq \mu_{x,i}(p_{x,i}-q_{x,i}) + \overline{v}_{r,i}.
\end{align}
which contradicts \eqref{eqn:dx_b2}. 
Therefore, the reference velocity vector $v_r$ in~\eqref{eqn:velcon} constrains the position error within the funnel, i.e.,
$$-\rho_x(t)\prec e_x(t)\prec \rho_x(t), \forall t\ge0.$$

\textbf{Stage II: Invariance of the $v$-funnel.} 
We again use contradiction to show that the control law~\eqref{eqn:con} keeps $e_v(t)$ inside $[-\rho_v(t),\rho_v(t)]$ as in \eqref{eqn:fun2}. 

Assume that $\tx$ is the first violation time. Then
\begin{align}\label{Eq:inqe_tv}
    &\exists i \in [1;n], e_{v,i}(\tx) \leq -\rho_{v,i}(\tx) \text{ or } e_{v,i}(\tx) \geq \rho_{v,i}(\tx) \notag \\
    &-\rho_{v,i}(t) < e_{v,i}(t) < \rho_{v,i}(t), \forall (t,i) \in [0, \tx) \times [1;n].
\end{align}
We will consider the following two cases for $t \in [0,\tx)$.

\begin{figure*}[t]
    \centering
    \includegraphics[width=\textwidth]{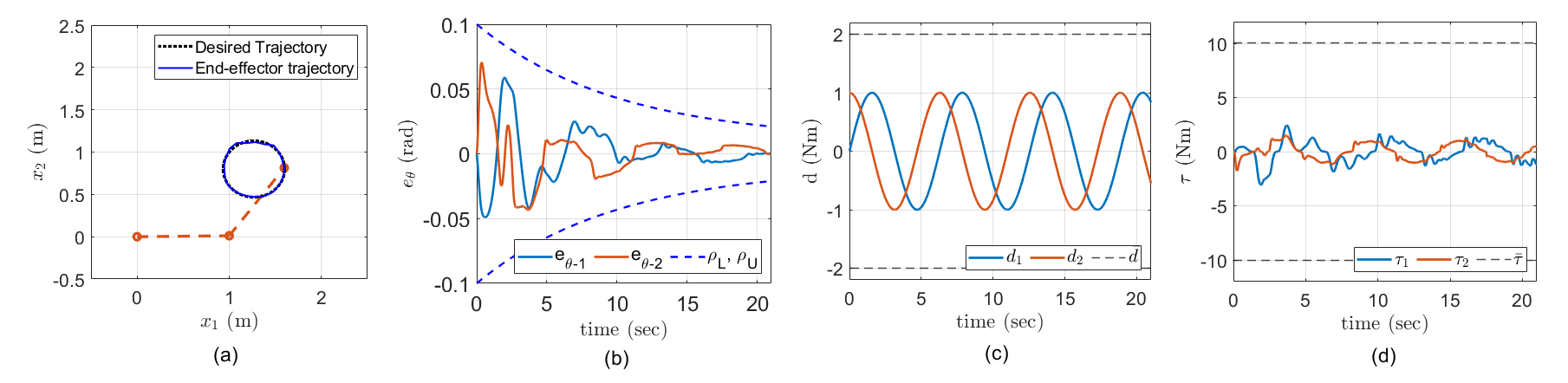}
    \caption{{Simulation results for 2R manipulator. (a) Desired trajectory vs tracked trajectory, (b) evolution of errors within funnel boundaries,  (c) sinusoidal disturbance with disturbance bounds, and (d) torque input with input bounds.}}
    \label{fig:sim_rr}
\end{figure*}

\textbf{Case I.} For some $i \in [1;n]$, $e_{v,i}(t)$ approaches the upper funnel constraint (Figure \ref{fig:case1}), i.e., $e_{v,i}(t) \rightarrow \rho_{v,i}(t) \implies e_{v,i}(t) - \rho_{v,i}(t) =: \overline{\alpha}_{v,i} \rightarrow 0$. Following \eqref{Eq:inqe_tv}, we have:
\begin{align*}
    &e_{v,i}(t) < \rho_{v,i}(t) \implies \overline{\alpha}_{v,i} \uparrow 0 \implies \lim_{\overline{\alpha}_{v,i} \uparrow 0} \frac{d}{dt} \overline{\alpha}_{v,i} > 0 \\
    \implies &\lim_{\overline{\alpha}_{v,i} \uparrow 0} \dot{e}_{v,i}(t) > \lim_{\overline{\alpha}_{v,i} \uparrow 0} \dot{\rho}_{v,i}(t) > -\mu_{v,i}(p_{v,i}-q_{v,i}) \\
    \implies &\lim_{\overline{\alpha}_{v,i} \uparrow 0} \dot{v}_{i}(t) > -\mu_{v,i}(p_{v,i}-q_{v,i}) + \dot{v}_r(t).
\end{align*}
Therefore, there exists $i \in [1;n]$, such that 
\begin{gather}\label{eqn:dv_b1}
    \lim_{\overline{\alpha}_{v,i} \uparrow 0} \dot{v}_{i}(t) > -\mu_{v,i}(p_{v,i}-q_{v,i}) - \overline{a}_{r,i}.
\end{gather}
Since, $\lim_{\overline{\alpha}_{v,i} \uparrow 0} \varepsilon_{v,i}(t) = 1$, we obtain $\lim_{\overline{\alpha}_{v,i} \uparrow 0} \tau_{i}(t) = -\overline{\tau}_{i}$.
Using the dynamics~\eqref{eqn:sysDyn_acc} and feasibility condition \eqref{eqn:feas2}
\begin{align}\label{eqn:dv_b1c}
    \lim_{\overline{\alpha}_{v,i} \uparrow 0} \!\!\dot{v}_{i}(t) \!\leq\! \overline{V}_{M,i} \!-\! \underline{m} \overline{\tau}_i \!+\! \underline{m}_i\overline{d}_i
    \!\leq\! \!-\mu_{v,i}(p_{v,i} \!-\! q_{v,i}) \!-\! \overline{a}_{r,i}
\end{align}
which contradicts \eqref{eqn:dv_b1}. 

\begin{figure*}[!t]
\centering
\includegraphics[width=\textwidth]{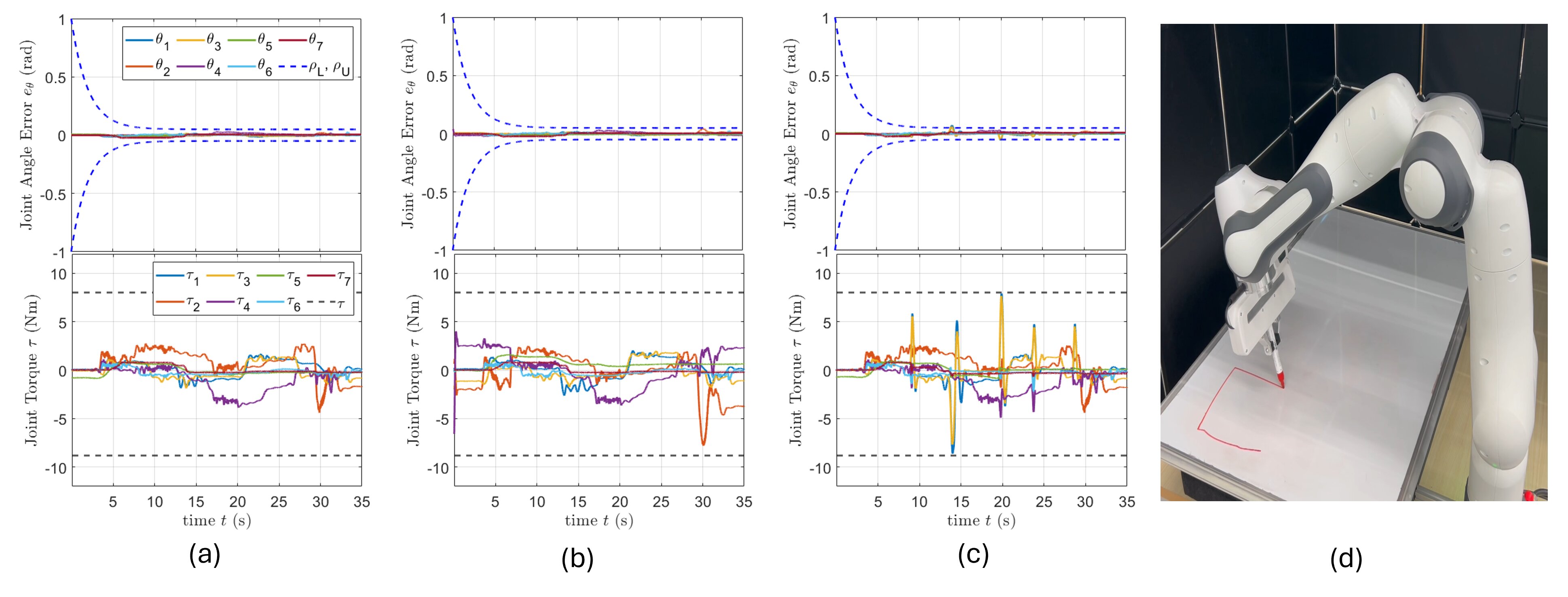}
\caption{7-DOF Franka Research 3 manipulator: tracking error constrained within funnels and torque input for (a) Nominal tracking, (b) Payload variation, and (c) External disturbances. \href{https://drive.google.com/file/d/1O5uEoPiUZU9zdcU34QYLhEC3ImSzORug/view?usp=sharing}{Video}.}
\label{fig:simFRANKA1}
\end{figure*}



\textbf{Case II.} For some $i \in [1;n]$, $e_{v,i}(t)$ approaches the lower funnel constraint (Figure \ref{fig:case2}), i.e., $e_{v,i}(t) \rightarrow -\rho_{v,i}(t) \implies e_{v,i}(t)+\rho_{v,i}(t) =: \underline{\alpha}_{v,i} \rightarrow 0$. Following \eqref{Eq:inqe_tx}, we have:
\begin{align*}
    &e_{v,i}(t) > -\rho_{v,i}(t) \implies \underline{\alpha}_{v,i} \downarrow 0 \implies \lim_{\underline{\alpha}_{v,i} \downarrow 0} \frac{d}{dt} \underline{\alpha}_{v,i} < 0 \\
    \implies &\lim_{\underline{\alpha}_{v,i} \downarrow 0} \dot{e}_{v,i}(t) < \lim_{\underline{\alpha}_{v,i} \downarrow 0} -\dot{\rho}_{v,i}(t) < \mu_{v,i}(p_{v,i}-q_{v,i}) \\
    \implies &\lim_{\underline{\alpha}_{v,i} \uparrow 0} \dot{v}_{i}(t) < \mu_{v,i}(p_{v,i}-q_{v,i}) + \dot{v}_r(t).
\end{align*}
Therefore, there exists $i \in [1;n]$, such that
\begin{gather}\label{eqn:dv_b2}
    \lim_{\underline{\alpha}_{v,i} \uparrow 0} \dot{v}_{i}(t) < \mu_{v,i}(p_{v,i}-q_{v,i}) + \overline{a}_{r,i}.
\end{gather}
Since, $\lim_{\underline{\alpha}_{v,i} \downarrow 0} \varepsilon_{v,i}(t) = -1$, we obtain $\lim_{\underline{\alpha}_{v,i} \downarrow 0} \tau_{i}(t) = \overline{\tau}_i.$ Using dynamics~\eqref{eqn:sysDyn_acc} and feasibility condition~\eqref{eqn:feas2}
\begin{align}\label{eqn:dv_b2c}
    \lim_{\underline{\alpha}_{v,i} \downarrow 0} \!\!\dot{v}_{i}(t) \!\geq\! \underline{V}_{M,i} \!+\! \underline{m} \overline{\tau}_i \!-\! \underline{m}_i\overline{d}_i \!\geq\! \mu_{v,i}(p_{v,i} \!-\! q_{v,i}) \!+\! \overline{a}_{r,i}
\end{align}
which contradicts \eqref{eqn:dv_b2}. 
Therefore, the controller $\tau$ in~\eqref{eqn:con} constrains the velocity error within the funnel, i.e.,
$$-\rho_x(t)\prec e_x(t)\prec \rho_x(t), \forall t\ge0.$$

Combining Stages~I and~II, the control laws~\eqref{eqn:velcon} and~\eqref{eqn:con} ensure funnel invariance and bounded signals, thereby guaranteeing tracking of the reference trajectory $x_{ref}(t)$.
\end{proof}

\begin{remark}
In Stage I, we do not assume perfect tracking of the virtual velocity, but consider the actual closed-loop dynamics $\dot{x}_i(t) = v_{r,i}(t) + e_{v,i}(t)$, where $e_v(t) = v(t) - v_r(t)$ is the velocity tracking error. Thus, Stage~I shows that $|e_x(t)| < \rho_x(t)$, provided that $|e_v(t)| < \rho_v(t)$ and \eqref{eqn:feas1} holds. Stage~II then shows that the controller~$\tau(t)$ in~\eqref{eqn:con} enforces the bound $|e_v(t)| < \rho_v(t)$, thereby ensuring tracking of $x_{ref}(t)$.
\end{remark}

\begin{remark}
    Assumptions~\ref{assum_d}-\ref{assum_Md} with known system parameter bounds (derived in practice using disturbance observers or parameter estimators) are only used to derive the feasibility conditions~\eqref{eqn:feas1}-\eqref{eqn:feas2}; the control law~\eqref{eqn:con} itself is model-free and does not require explicit knowledge of the system dynamics. When these feasibility conditions hold, Theorem~\ref{thm:bdcontrol} guarantees funnel invariance and bounded inputs. 
    
    If feasibility is violated, prescribed performance cannot be ensured; instead, the transformations keep the input bounded and promote safe behavior. With the saturation transform, the controller applies the maximum allowable input in an attempt to recover, while the zeroing transform gradually reduces the control effort to prevent further deviation.
\end{remark}

\section{Simulation and Experimental Results}\label{sec:sim}
We now demonstrate the proposed controller on three platforms: (i) a two-link manipulator, (ii) a 7-DOF manipulator, and (iii) an omnidirectional mobile robot. We also compare the two bounded transformation functions and evaluate the performance against baseline approaches.
\begin{figure*}[t]
    \centering
    \includegraphics[width=\textwidth]{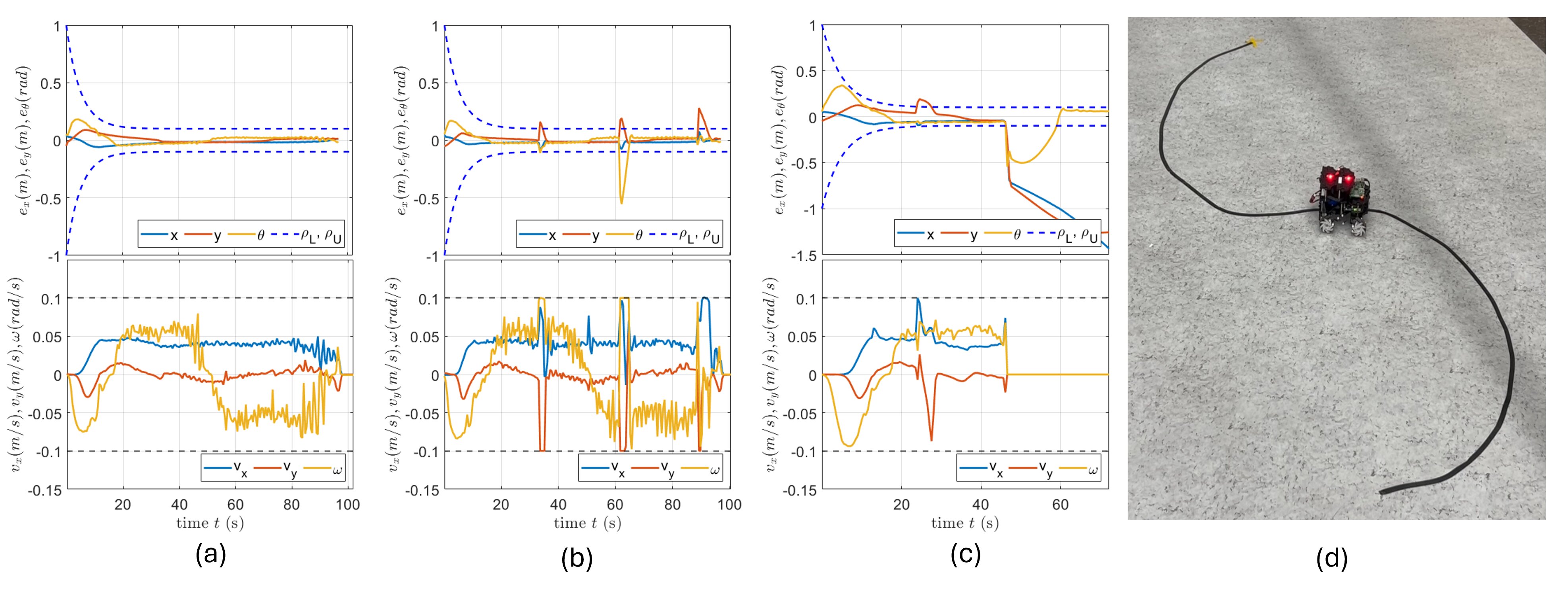}
    \caption{Mobile-robot: (a) Nominal tracking, (b), (c) Disturbance rejection with Saturation and Zeroing transformations. \href{https://drive.google.com/file/d/1DLhvGYIsafk7bTTYYLXRbRcYz5F-ggzz/view?usp=sharing}{Video}}
    \label{fig:simomni}
\end{figure*}

\subsection{Two-link SCARA manipulator}
We first consider a two-link SCARA manipulator adapted from \cite{spong}:
\begin{gather}\label{eqn:rr}
    ml^2
    \begin{bmatrix}
        \frac{5}{3} + c_2 & \frac{1}{3} + \frac{1}{2}c_2 \\
        \frac{1}{3} + \frac{1}{2}c_2 & \frac{1}{3}
    \end{bmatrix}
    \begin{bmatrix}
        \Ddot{\theta}_1 \\
        \Ddot{\theta}_2
    \end{bmatrix} 
    +
    ml^2s_2
    \begin{bmatrix}
        -\frac{1}{2}\dot{\theta}_2^2 - \dot{\theta}_1\dot{\theta}_2\\
        \frac{1}{2}\dot{\theta}_1^2
    \end{bmatrix}
    + mgl
    \begin{bmatrix}
        \frac{3}{2}c_1 + \frac{1}{2}c_{12} \\
        \frac{1}{2}c_{12}
    \end{bmatrix}
    = 
    \begin{bmatrix}
        \tau_1(t) \\
        \tau_2(t)
    \end{bmatrix}
    + d(t). \nonumber
\end{gather}
The link masses and lengths, gravitational effects, and external disturbance terms are assumed to be unknown. The velocity and torque limits are set to $\overline{v}=6$~rad/s and $\overline{\tau}=10$~N-m and the funnel parameters are chosen as $p_x=0.2$, $q_x=0.02$, $\mu_x=0.1$ and $p_v=2$, $q_v=0.02$, $\mu_v=0.1$.
We define the parameter bounds in Assumptions \ref{assum_tau}-\ref{assum_Md} as $\underline{m}=1.5 \text{ kg}, \underline{m}_i = 1.6 \text{ kg}^{-1}$, and $\max{(-\underline{V}_M, \overline{V}_M)} = 5 m^2/s^2$. From the feasibility conditions~\eqref{eqn:feas1} and \eqref{eqn:feas2}, we determine the maximum permissible disturbance as $\overline{d} = 2 N-m$. 
As shown in Figure~\ref{fig:sim_rr}, the controller keeps the tracking error inside the funnels while respecting input limits.

\subsection{7-DOF Franka Research 3 Manipulator}
The second case study uses a 7-DOF Franka Research~3 robot \cite{FRANKA} with joint limits $\overline{v}=6$~rad/s and $\overline{\tau}=8$~N-m. 
Funnel parameters are set to $p_x=1$, $q_x=0.1$, $\mu_x=0.5$ and
$p_v=2$, $q_v=0.1$, $\mu_v=0.5$ for all the 7 joints.
Three experiments are performed: (a) nominal tracking, (b) payload variation (attaching a water bottle to one of the links) to illustrate robustness to unknown dynamics, and (c) external disturbances. In all cases, the controller maintains accurate tracking while satisfying input constraints. Hardware results are shown in Figure~\ref{fig:simFRANKA1}, with videos available \href{https://drive.google.com/file/d/15S_USJ11sCcmMG9JdmNWkoCA4rXxKiZf/view?usp=sharing}{here}.

\subsection{Mobile Robot}
The third case study validates the method on an omnidirectional mobile robot with dynamics from~\cite{STT}:
\begin{align}
    \begin{bmatrix}
        \dot{x} \\ \dot{y} \\ \dot{\theta}
    \end{bmatrix}
    = 
    \begin{bmatrix}
        \cos{\theta} & \sin{\theta} & 0 \\ \sin{\theta} & -\cos{\theta} & 0 \\ 0 & 0 & 1
    \end{bmatrix}
    \begin{bmatrix}
        v_x \\ v_y \\ \omega
    \end{bmatrix} + d(t).
\end{align}
The linear and angular velocity limits are set to $0.1$~m/s and $0.1$~rad/s, respectively.
When feasibility conditions hold, the tracking error remains inside the
funnels (Figure~\ref{fig:simomni}(a)). When sudden disturbances violate
feasibility, Figures~\ref{fig:simomni}(b) and \ref{fig:simomni}(c) shows the effect of saturation and zeroing transformations. Hardware demonstration videos can be found in this \href{https://drive.google.com/file/d/1DLhvGYIsafk7bTTYYLXRbRcYz5F-ggzz/view?usp=sharing}{Link}.

\begin{figure*}[t]
    \centering
    \includegraphics[width=\textwidth]{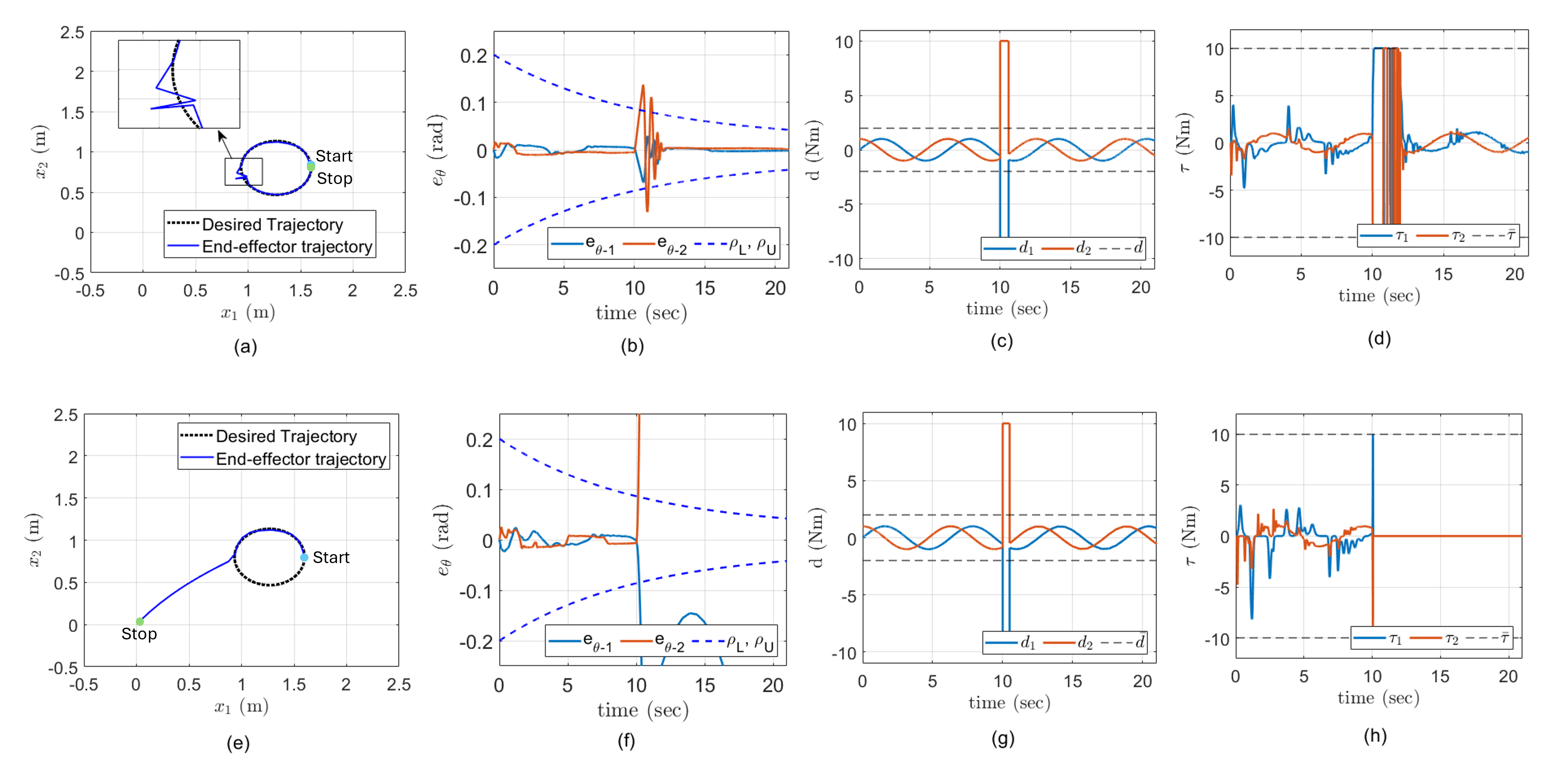}
    \caption{Bounded Transformation Function Comparison.}
    \label{fig:rr_comp}
\end{figure*}

\begin{table}[t]
\centering
\caption{Quantitative comparison of tracking performance for nominal and disturbed cases.}
\label{tab:comparison}
\setlength{\tabcolsep}{5pt}
\renewcommand{\arraystretch}{1.15}
\resizebox{0.5\columnwidth}{!}{%
\begin{tabular}{lcccccc}
\toprule
\textbf{Method} &
\multicolumn{3}{c}{\textbf{Nominal Case}} &
\multicolumn{3}{c}{\textbf{Disturbed Case}} \\
\cmidrule(lr){2-4} \cmidrule(lr){5-7}
& \textbf{Error} & $\|\mathbf{u}\|_{\max}$ & \textbf{Time (s)}
& \textbf{Error} & $\|\mathbf{u}\|_{\max}$ & \textbf{Time (s)} \\
\midrule
Proposed & 
$\mathbf{0.018}$ & $\mathbf{1.407}$ & $\mathbf{0.013}$ &
$\mathbf{0.036}$ & $\mathbf{1.142}$ & $\mathbf{0.015}$ \\

PPC & 
$0.019$ & $1.136$ & $0.012$ &
$0.074$ & $5.511$ & $0.011$ \\

PID & 
$0.052$ & $1.047$ & $0.011$ &
$0.497$ & $2.408$ & $0.010$ \\

MPC & 
$0.098$ & $1.032$ & $1.146$ &
$1.907$ & $0.953$ & $1.092$ \\
\bottomrule
\end{tabular}}
\end{table}

\subsection{{Comparison of the bounded transformation functions}}
The effectiveness of the proposed control framework is further evaluated by comparing the two types of bounded transformation functions, saturation and zeroing. These transformations determine the system's response when the feasibility conditions are violated, each offering distinct advantages based on task requirements. This evaluation is conducted through two case studies: a 2R manipulator, analyzed in simulation, and a mobile robot, tested experimentally in hardware, as shown in Figures \ref{fig:rr_comp} and \ref{fig:simomni}, respectively. The simulation facilitates precise computation of system parameter bounds and disturbance limits to validate feasibility conditions in Equations \eqref{eqn:feas1} and \eqref{eqn:feas2}, while the hardware demonstration displays the framework's practicality and robustness in real-world scenarios.

The saturation transformation keeps the control input active within the input bounds, allowing the system to recover and resume trajectory tracking once disturbances subside. For the 2R manipulator (Figure \ref{fig:rr_comp}(a)), a sudden jerk disrupts (Figure \ref{fig:rr_comp}(c)) the tracking of the circular trajectory. However, the control input (Figure \ref{fig:rr_comp}(d)) remains within the allowable bounds, ultimately driving the tracking error back within the funnel bounds, as seen in Figure \ref{fig:rr_comp}(b). Similarly, in the mobile robot case, when the tracking error exceeds the funnel bounds due to sudden jerks (Figure \ref{fig:simomni}(b)), the control input (Figure \ref{fig:simomni}(f)) stays within bounds, and successfully guides the tracking error back inside the funnel. This approach favors task continuity, making it suitable for applications like assembly-line operations or precise path-following in less hazardous environments.

In contrast, the zeroing transformation halts the system by driving the control input to zero whenever the tracking error exceeds the funnel bounds due to the violation of the feasibility conditions. In the 2R manipulator case (Figure \ref{fig:rr_comp}(e)), a sudden jerk causes the system to fail in tracking the circular trajectory (Figure \ref{fig:rr_comp}(g)). Consequently, the control input (Figure \ref{fig:rr_comp}(h)) drops to zero, halting the manipulator. Similarly, in the mobile robot case, when a sudden jerk causes the tracking error to exceed the funnel bounds (Figure \ref{fig:simomni}(c)), the control input (Figure \ref{fig:simomni}(f)) also drops to zero, bringing the robot to a stop. This approach is ideal for safety-critical tasks, such as industrial manipulators operating near humans or fragile equipment and mobile robots navigating in hazardous or unknown terrain.

\subsection{Comparison with Baseline Algorithms}
Table~\ref{tab:comparison} compares the proposed controller with PPC~\cite{PPC1}, PID, and MPC~\cite{MPC2} for trajectory tracking under nominal and disturbed conditions. All simulations were carried out on a mobile robot model using an Intel i9-10900 @ 2.80\,GHz processor with 16\,GB RAM. The metrics include RMS tracking error, input magnitude, and average computation time. In the nominal case, the proposed method achieves performance comparable to PPC and MPC while maintaining bounded inputs, whereas MPC incurs significantly higher computation time due to online optimization. In the disturbed case, the benefits of the bounded transformation functions become more evident. PPC and PID fail to respect input constraints under large disturbances, while MPC maintains bounded inputs but exhibits degraded tracking performance due to its reliance on accurate system information. These results highlight the effectiveness of the proposed approach for real-time applications with input constraints.

\section{Conclusion}
This paper presented a prescribed performance control framework for unknown Euler–Lagrange systems with input constraints. The proposed controller is approximation-free, computationally efficient, and enforces hard funnel constraints while guaranteeing bounded control inputs through explicit feasibility conditions. Two bounded transformation functions, saturation and zeroing, were introduced, which allow the controller to prioritize either task continuity or safety. Simulation and hardware experiments demonstrated reliable tracking performance, robustness to disturbances, and real-time applicability. 

\bibliographystyle{unsrt} 
\bibliography{sources}

\end{document}